\documentclass[a4paper]{ifacconf}

\usepackage[utf8]{inputenc}

\usepackage{graphicx}      
\usepackage{natbib}        
\usepackage{glossaries-prefix}
\newacronym{pid}{PID}{proportional-integral-derivative}
\newacronym[prefixfirst={a\ },
            prefix={an\ }
            ]{ml}{ML}{machine learning}
\newacronym{drl}{DRL}{deep reinforcement learning}
\newacronym{mlp}{MLP}{multi layer perceptron}
\newacronym[prefixfirst={a\ },
            prefix={an\ }
            ]{nn}{NN}{neural network}
\newacronym[prefixfirst={a\ },
            prefix={an\ }
            ]{rl}{RL}{reinforcement learning}
\newacronym{bc}{BC}{behaviour cloning}
\newacronym[prefixfirst={a\ },
            prefix={an\ }
            ]{mdp}{MDP}{Markov decision process}
\newacronym{ddpg}{DDPG}{deep deterministic policy gradients}
\newacronym{td3}{TD3}{twin delayed DDPG}
\newacronym{her}{HER}{hindsight experience replay}
\newacronym{roa}{RoA}{region of attraction}
\newacronym{dqfd}{DQfD}{deep q-learning from demonstrations}
\newacronym{ddpgfd}{DDPGfD}{deep deterministic policy gradients from demonstrations}
\newacronym{asddpg}{AsDDPG}{assisted deterministic policy gradients}
\newacronym{il}{IL}{imitation learning}
\usepackage{amsmath,amssymb,amsfonts}
\usepackage{algorithmic}
\usepackage{graphicx}
\usepackage{textcomp}
\usepackage{bbold}
\usepackage{subcaption}
\RequirePackage[absolute]{textpos}
\DeclareRobustCommand*{\copyrightnote}[1]{%
  \begin{textblock}{150}(50,290)
      \textbf{#1}%
  \end{textblock}%
    }

\begin{document}
\setlength{\abovedisplayskip}{-7.5pt}
\setlength{\abovedisplayshortskip}{-7.5pt}
\begin{frontmatter}

\title{Accelerating Reinforcement Learning with Suboptimal Guidance} 

\thanks[footnoteinfo]{The first author is financed by "PhD Scholarships at SINTEF" from the Research Council of Norway (grant no. 272402). The third author was supported by the Research Council of Norway (grant no. 223254 NTNU AMOS).}

\author[First]{Eivind Bøhn} 
\author[First,Second]{Signe Moe} 
\author[Second]{Tor Arne Johansen}

\address[First]{SINTEF Digital, Oslo, Norway\\ (e-mail: \{eivind.bohn, signe.moe\}@sintef.no).}
\address[Second]{Norwegian University of Science and Technology, Trondheim, Norway (email: \{signe.moe, tor.arne.johansen\}@ntnu.no)}


\begin{abstract}
Reinforcement Learning in domains with sparse rewards is a difficult problem, and a large part of the training process is often spent searching the state space in a more or less random fashion for any learning signals. For control problems, we often have some controller readily available which might be suboptimal but nevertheless solves the problem to some degree. This controller can be used to guide the initial exploration phase of the learning controller towards reward yielding states, reducing the time before refinement of a viable policy can be initiated. In our work, the agent is guided through an auxiliary behaviour cloning loss which is made conditional on a Q-filter, i.e. it is only applied in situations where the critic deems the guiding controller to be better than the agent. The Q-filter provides a natural way to adjust the guidance throughout the training process, allowing the agent to exceed the guiding controller in a manner that is adaptive to the task at hand and the proficiency of the guiding controller. The contribution of this paper lies in identifying shortcomings in previously proposed implementations of the Q-filter concept, and in suggesting some ways these issues can be mitigated. These modifications are tested on the OpenAI Gym Fetch environments, showing clear improvements in adaptivity and yielding increased performance in all robotic environments tested. 
\end{abstract}

\begin{keyword}
Deep Reinforcement Learning, Non-Linear Control Systems, Robotics
\end{keyword}

\end{frontmatter}


\section{INTRODUCTION}\label{sec:introduction}
\Gls{rl}~\citep{sutton_reinforcement_nodate} is a field of machine learning concerned with sequential decision making problems. The theory and methods in \gls{rl} has produced some impressive results the last few years, ranging from high-level reasoning tasks such as game-playing to control of fast dynamics such as actuation in robotics. \gls{rl} has received much attention and interest due to its framework being very general, in theory capable of solving nearly any problem as long as one can define some notion of utility of different states of the system.
\copyrightnote{This work has been submitted to IFAC for possible publication}

This utility measure, called the reward function in the \gls{rl} framework, is central to both the definition of the problem and the performance of the algorithm. Often, a sparse reward function\footnote{A sparse reward function is one that only yield signals in some subset of the state space, e.g. in the subset defined as the set of goal states, and is typically constant elsewhere.} is preferable, as they are easier to formulate, easier to estimate, and more importantly, there is less room for ambiguity and misinterpreting the objective of the task. The obvious downside is however that only a small region of the problem space actually gives a learning signal to the agent, and a significant portion of the training time is spent exploring the state space (often in a random manner) until a minimally viable strategy (i.e. one that consistently is able to reach reward-yielding states) is found, from which further progress can be made. 

Often in robotics and other control applications one already has a controller, which due to e.g. nonlinearities in the dynamics, uncertainties in model parameters, or strict computational requirements might be suboptimal. This controller could be used to guide the learning controller towards this minimally viable strategy, thereby reducing the long initial exploration phase. In the rest of this paper guiding controller and pre-existing controller is used interchangeably to refer to this controller.

The most naive implementation of \gls{il}, that is, trying to directly copy the pre-existing controller is seldom successful, mainly due to a data distribution mismatch. Furthermore, the potential performance of the learning controller is bound by the performance of the existing controller. The data distribution problem arises as the data collected by the demonstrator will only consist of a subset of the state space, and will offer no guidance on how to course correct back into this subset when the learning controller inevitably makes a small deviation from the controller it is imitating. Small errors therefore tend to accumulate into trajectories that stray far from those produced by the demonstrator. Additionally, when combining \gls{il} with \gls{rl} there needs to be some mechanism in place to allow the learning controller to make different choices than the original controller when appropriate, in order to exceed the performance of the existing controller.

The problem of leveraging information encoded in previous controllers or demonstration data and combining it with further learning such as \gls{rl} has been widely studied. This has led to the development of several different approaches, such as inverse reinforcement learning, \gls{bc}, and pretraining learning controllers with demonstration data. In our work, we tackle the problem of deciding when the learning controller should imitate the pre-existing controller, and when it should develop its own behaviour. We base our method in the concept of the Q-filter~\citep{nair_overcoming_2017}, which makes the learning controller imitate the pre-existing controller only when it is deemed to yield a higher reward than what the learning controller would achieve. In this work we identify issues with the original formulation of the Q-filter and propose modifications to mitigate these issues, resulting in improved adaptivity to the difficulty of the task and to the proficiency of the pre-existing controller, and thus also improved performance.

The rest of the paper is organized as follows. First, related work is presented and discussed in Section \ref{sec:related}, then the requisite background theory for \gls{rl} and the algorithms used in this paper is presented in Section \ref{sec:background}. Our method is outlined in Section \ref{sec:method}, and the experiments used to evaluate the methods are outlined in Section \ref{sec:experiments}. The results of the experiments are shown in Section \ref{sec:discussion}, which also discusses the strengths and weaknesses of the proposed method. Finally, Section \ref{sec:conclusion} concludes with our thoughts on the matter and suggestions for further work.
\section{RELATED WORK}\label{sec:related}
The DAGGER algorithm~\citep{ross_reduction_nodate} is an approach to mitigating the data distribution mismatch problem. A learning controller is trained on a dataset of experience, and then this learning controller is used in the environment to collect more data. The newly collected data is appended to the training set, and the demonstrator is asked to label the new data with the appropriate outputs. Our method builds upon the DAGGER framework in the sense that we assume access to an online demonstrator, and iteratively expand the dataset with the demonstrators knowledge. However, our learning controller is only made to mimic the demonstrator in some selected states and this is only one of its optimization objectives. Deeply AggreVaTeD~\citep{sun_deeply_2017} is a further extension of DAGGER to continuous state and action spaces and nonlinear function approximation with \glspl{nn}. Both these approaches are pure \gls{il} methods and thus do not allow for the agent to surpass the demonstrator.

\Gls{ddpgfd}~\citep{vecerik_leveraging_2018} is a method for leveraging demonstration data for \gls{rl} algorithms in domains with continuous state and action spaces. In this work, human generated demonstration are included in the replay buffer for which a prioritized sampling technique is used to ensure a suitable mix of demonstration data and self collected data in each training batch. They further use a mix of 1-step and n-step return losses for training the Q-networks. They show increased performance over the demonstrations, as well as over regular \gls{ddpg} even when the latter uses hand-crafted shaped rewards and \gls{ddpgfd} uses sparse rewards. 

\cite{nair_overcoming_2017} propose to address the problem of choosing when the learning controller should emulate the demonstrator with the use of the Q-filter. The Q-filter is an indicator function used to select when to apply a behaviour cloning loss on the action chosen by the demonstrator. The filter evaluates to true when the estimated value from the Q-function is higher for the demonstrator action than it is for the action chosen by the learning controller. This provides a natural way to anneal behaviour cloning loss during the training process that is adaptive to the task at hand. Their demonstrations come in the form of a fixed set of trajectories collected by a human demonstrator in a virtual reality version of the environment. In each training batch, a portion of the data is sampled from the regular replay buffer and a portion is sampled from the demonstration buffer, on which \gls{bc} is applied for the actions selected by the Q-filter. Our method borrows the concept of the Q-filter from this work, suggesting some important adjustments to the concept. Further, their source of demonstrations consists of a fixed dataset, while we assume online access to a demonstrator controller.

In \gls{asddpg}~\cite{xie_learning_2018} propose a novel architecture for incorporating a simple controller to guide the initial exploration phase.  In their architecture, the critic module has an additional branch that estimates the Q-values of the pre-existing controllers action and the actors action for the state in question. Based on these estimates the method greedily chooses the one with the highest Q-value, and applies this action as the exploration action during training. In this way, the guiding controller is used to show the actor some primitives it can use to aid exploration. As in our method, online access to the guiding controller is assumed, but the Q-filter is applied to select actions for exploration instead of to shape the gradients of the optimization procedure.


\section{BACKGROUND}\label{sec:background}
\subsection{Reinforcement Learning}
We consider the \gls{rl} problem in a \gls{mdp} framework. The \gls{mdp} is defined by the tuple $\langle \mathcal{S}, \mathcal{A}, R, \mathcal{T}, \gamma \rangle$, where $\mathcal{S}$ is the set of states, $\mathcal{A}$ is the set of actions, $R(s, a)$ is the reward function, $\mathcal{T}$ is the state transition function describing the evolution of the states as a function of time and actions, and $\gamma \in [0, 1]$ is the discount factor, weighing the relative importance of immediate and future rewards. We denote a trajectory $\tau$ as a sequence of state and actions, e.g. one episode of the problem task, and the return as $R(\tau) = \sum_{(s_t, a_t) \sim \tau} \gamma^t R(s_t, a_t)$, where $\sim$ signifies that the left hand side is distributed according to the distribution of the right hand side. The \gls{rl} objective is to find a policy $\pi$, i.e. a function that generates actions from states, which maximizes the sum of expected future rewards weighed by the discount factor $\gamma$, i.e. the expected return in an episodic setting. We denote this objective as $\max_\theta J(\theta) = \mathbb{E}_{\tau \sim \pi_\theta} \left[ R(\tau) \right]$, where $\theta$ is the parameterization of the policy. Central to many \gls{rl} algorithm is the concept of the value of a state, given by the value function $V^\pi(s) = \mathbb{E}_{\tau \sim \pi} \left[R(\tau) \enspace | \enspace s_0 = s \right]$. The value function measures the expected return of being in a given state and from there always acting according to the policy $\pi$. If we further leave the first action in the trajectory free, we get the state-action value function, often called the Q-function $Q^\pi(s, a) = \mathbb{E}_{\tau \sim \pi} \left[R(\tau) \enspace | \enspace s_0 = s, \enspace a_0 = a \right]$. We consider in this work a further extension to the \gls{mdp} framework in the form of a specified goal state, from a set of possible goals $\mathcal{G}$. \cite{schaul_universal_nodate} show that value functions conditioned on this additional goal parameter can successfully be trained to generalize to unseen goals. 

\subsection{Deep Deterministic Policy Gradients}
To address the challenges posed by tasks with continuous state and action spaces, \citet{lillicrap_continuous_2015} introduced \gls{ddpg}. \gls{ddpg} is an off-policy, model-free, policy-gradient, actor-critic algorithm which concurrently estimates the Q-function $Q^{\pi_\theta}$ and an actor that aims at solving for the actions which maximize the Q-function. These actor and critic functions are implemented as \glspl{nn}, and are trained off-policy from experience collected by the actor interacting with the environment. This experience is stored in a replay buffer $\mathcal{D}$, from which we sample minibatches $\mathcal{B}$ to train the actor and the critic. More concretely, the critic is trained in a supervised manner to regress on 1-step returns according to the Bellman optimality equation:

\begin{align}
    y_t &= R_t + \gamma Q(s_{t+1}, \pi(s_{t+1} | \theta^Q)) \label{eq:q-target} \\ 
    \mathcal{L}(\theta^Q) &= \mathbb{E}_{(s_t, a_t, R_t, s_{t+1}) \sim \mathcal{B} } \left[(y_t - Q(s_t, a_t | \theta^Q))^2 \right]
\end{align}

where $\theta^Q$ is the parameterization of the Q-function. The actor attempts to find the optimal policy according to the critic through the relationship \eqref{eq:policy_from_critic}. Since this actor is deterministic, a stochastic behaviour policy \eqref{eq:behaviour} is used to collect experience in order to improve exploration, which is obtained by adding Gaussian noise with parameters selected to suit the problem. Finally, the actor is trained by optimizing the objective \eqref{eq:ddpg_objective}.

\begin{align}
    \pi(s) &= \textrm{arg}\max_a Q(s, a) \label{eq:policy_from_critic} \\
    \pi^b(s) &= \pi(s) + \mathcal{N}(\mu, \sigma) \label{eq:behaviour} \\
    \mathcal{L}(\theta) &= \max_\theta \mathbb{E}_{s \sim \mathcal{D}} \left[ Q_{\theta^Q} (s, \pi_\theta(s)) \right] \label{eq:ddpg_objective}
\end{align}

\subsection{Twin Delayed Deep Deterministic Policy Gradients}
\Gls{td3}~\citep{fujimoto_addressing_2018} improves upon \gls{ddpg} in several ways, notably by alleviating sources of overestimation in the critic network, and introduces more robustness towards hyperparameters. It does this by maintaining two separate networks estimating the critic Q-function, and using the lesser of these two as the target for regression \eqref{eq:q-target}. Further they suggest updating the policy networks less frequently to allow the value functions more time to converge before they are used to update the policy. 

\subsection{Hindsight Experience Replay}
The environments considered in this paper have sparse reward structures, requiring a directed effort over several actions to reach regions of the state space where any learning signal is received at all. To reach learning signals in such scenarios more quickly, \cite{andrychowicz_hindsight_2017} proposed \gls{her}, in which the experience contained in the replay buffer is retroactively fitted with a new goal state that was actually achieved sometime during the episode. In this way, by pretending that some state the actor was successful in achieving was the actual goal state, the algorithm receives more frequent learning signals and convergence is accelerated.

\section{METHOD}\label{sec:method}
\subsection{Problem Description}
In our setup we assume online access to a pre-existing sub-optimal controller which is capable of achieving goal states from a non-zero share of initial states. We also assume that we have access to the Q-function of this controller, denoted by $Q^G$. However, we do not assume prior access to demonstration trajectories from this controller. 

\subsection{Proposed Method}
We modify the actor objective function of \gls{ddpg} style algorithms \eqref{eq:ddpg_objective} by adding a \gls{bc} loss term, and as in~\cite{nair_overcoming_2017} we only selectively apply this loss with the indicator function conditioned on the Q-filter:

\begin{align}\label{eq:bc_loss_term}
    \mathcal{L}_{BC} &= \sum_{i=0}^{N_b} \left\Vert a_i - \pi_\theta(s_i) \right\Vert_2 \mathbb{1}_{Q^{G}(s_i, a_i) > Q^{\pi_\theta} (s_i, \pi_{\theta}(s_i))} \\
    &\textrm{where }\mathbb{1}_{A > B} = 
    \begin{cases}
    1 \quad\textrm{if } A > B \\
    0 \quad\textrm{otherwise}
    \end{cases} \\
    Q^{\pi_{\theta_0}} &\leftarrow Q^G \label{eq:q_init}
\end{align}

Here $N_b$ is the number of samples in each training batch, i.e. we use the online availability of the pre-existing controller to apply this term to all samples in the batch. This does not add considerable overhead, as the pre-existing controllers action can be evaluated once and then saved to the replay buffer alongside the other data.\footnote{Training with a guiding controller increases the wall clock time of the training process by about 5\% compared with normal training in our experiments.} In this work we suggest to modify the original formulation of the Q-filter by replacing the left hand side of the condition with the pre-existing controllers own estimated Q-function, which is kept static throughout the training process. As the \gls{td3} algorithm has two separate Q-networks, one can look at different ways of combining these to make up the Q-filter. In this work we have chosen to only use the Q-network that is used in the optimization objective \eqref{eq:ddpg_objective} as a basis for the comparison.

In order for the actors and the guiding controllers Q-functions to have comparable magnitudes, the actors Q-network is initialized to that of the guiding controller \eqref{eq:q_init}. This also provides a starting point that should be closer to optimal than a random initialization.

The motivation behind this modification is based on two key insights about the formulation of the original Q-filter in~\cite{nair_overcoming_2017}: First, for their Q-filter to have its intended effects the Q-network need to have converged, otherwise the comparison of actions will be highly stochastic. Moreover, the action frequency in control applications is typically high, thereby only to a small degree affecting the subsequent state, thus the sides of the inequality will differ principally in the immediate reward, which in most cases will also be similar. The quantities that are compared will thus be very similar and the action discrimination is therefore further stochastisized. Second, the actor in \gls{ddpg} style algorithms directly implements a maximization over actions of the Q-function, and will therefore quickly learn to optimize the unconverged Q-network, therefore seemingly (most times erroneously) offering better options than the pre-existing controller. Their implementation will henceforth be referred to as the naive method.

In practice, the Q-filter therefore tends to prefer the guiding controllers actions infrequently in the beginning, and converge to imitate the guiding controller with some non-zero frequency. We want the learning controller to aim to emulate the pre-existing controller to a large degree in the beginning, and then have the \gls{bc} loss taper off to allow the learning controller to surpass the pre-existing controller. With the original formulation of the Q-filter however, the \gls{bc} loss does not seem to be strategically applied. We therefore suggest using a different Q-function to assess the value of the pre-existing controllers actions, which is found by fitting a Q-network to data gathered by this controllers actions. This Q-function provides a more reliable estimate of the guiding controller actions true value right from the start, and also provides a static goal to compare against as the training progresses. 

\subsection{Guiding Controllers}
The guiding controllers are handcrafted proportional controllers with some conditional logic, e.g. move over block, then lower hand and grip block etc. The controllers exhibit varying proficiency levels, from a perfect\footnote{Perfect in the sense that it is able to complete all tasks in the test set.} controller capable of achieving any goal in the test set for the pick-and-place environment, to a controller achieving merely 21\% of the goals in the slide task.

\section{EXPERIMENTS}\label{sec:experiments}
We train and evaluate our method on the OpenAI Gym Fetch environments~\citep{plappert_multi-goal_2018}, which are based on the MuJoCo physics simulator~\citep{mujoco}. The FetchReach environment is too easy to solve for the \gls{rl} algorithms even without the use of a guiding controller, and as such has been excluded from the experiments. For each environment a test set consisting of 100 random initial states and goal states has been constructed, and the agents are evaluated on this set every 20k time steps. As a baseline, we evaluate our method against the original formulation of the Q-filter, as well as against an approach where the weight of the \gls{bc} loss is linearly decayed with time steps. We run each experiment with five different random seeds and show mean results with one standard deviation confidence bounds.

\subsection{Configuration}
Due to limited computational resources we employed the \gls{td3} algorithm instead of \gls{ddpg}, as convergence of the latter is often dependent on running several actors in parallel. Following the original paper introducing the Fetch environments \citep{andrychowicz_hindsight_2017}, we use a decay factor of $0.98$ and clip the regression Q-targets \eqref{eq:q-target} to the ranges possible in the environment, and also continue bootstrapping on environment termination due to time limit. We use the hyperparameters from \gls{td3}, i.e. two fully-connected hidden layers of 400 and 300 nodes respectively, a learning rate of 0.001 for both actor and critic, and a replay buffer size of $10^6$. Training is run every 1000 time steps of the environment for 1000 gradient steps, with a batch size of 100. We apply a small $\mathcal{L}_2$ regularization term with strength $0.01$ to the pre-activation values of the actors output layer, in order to mitigate vanishing gradients issues. The \gls{bc} loss term \eqref{eq:bc_loss_term} is weighted with a factor of 2. 

The linear schedule is set to be fully decayed after 500k time steps, with an initial value equal to that of the other methods. This schedule is not optimal for all environments, but a static schedule was chosen to highlight some of the issues with this approach.

\subsection{Ablation Studies}
It is conceivable that the improvements of the proposed methods stem mainly from the knowledge encoded into $Q^G$ and the initialization of $Q^{\pi_\theta}$ to this. A pretrained $Q^G$ already contains information about which action would maximize the rewards when further following the same action choices as the pre-existing controller at each state, and the optimization procedure implemented by the actor could take advantage of this information to achieve some of the same benefits that the \gls{bc} loss offers. To test this hypothesis, we ran one version of our method without the \gls{bc} loss term, and one version of the naive method which was initialized to $Q^G$ as in \eqref{eq:q_init}.

\section{RESULTS AND DISCUSSION}\label{sec:discussion}
\subsection{Results of Experiments}
\begin{figure*}[thpb]
    \centering
    \begin{subfigure}[b]{\linewidth}
        \centering
        \includegraphics[scale=0.5]{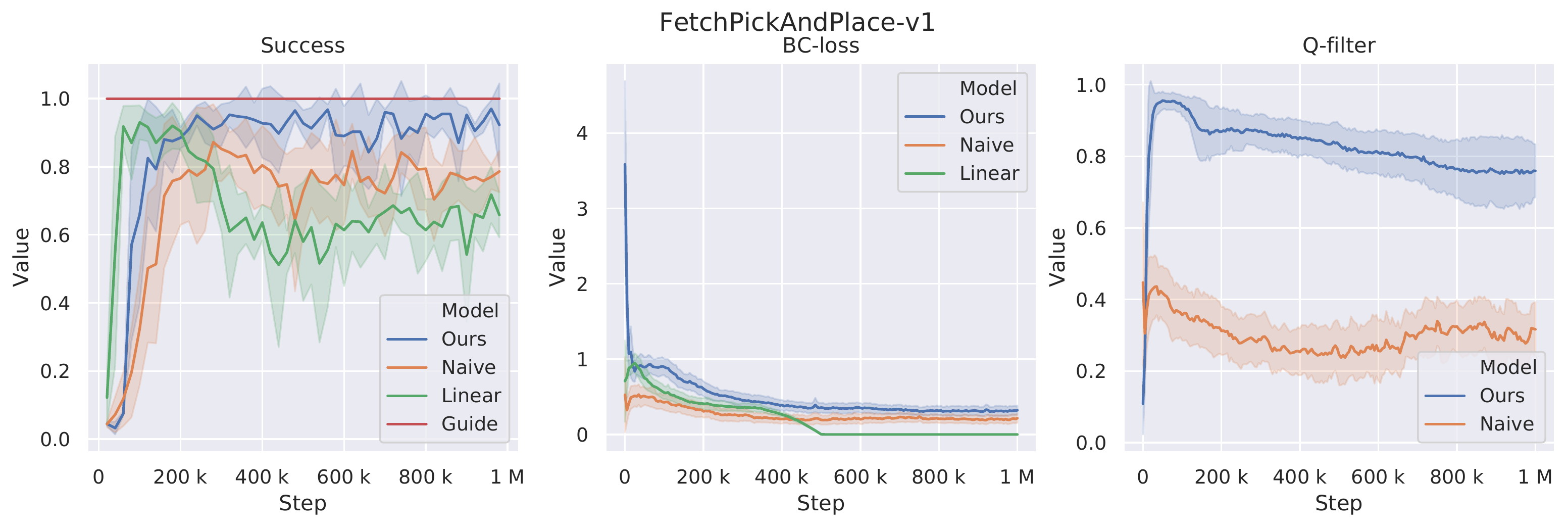}
    \end{subfigure}
    \begin{subfigure}[b]{\linewidth}
        \centering
        \includegraphics[scale=0.5]{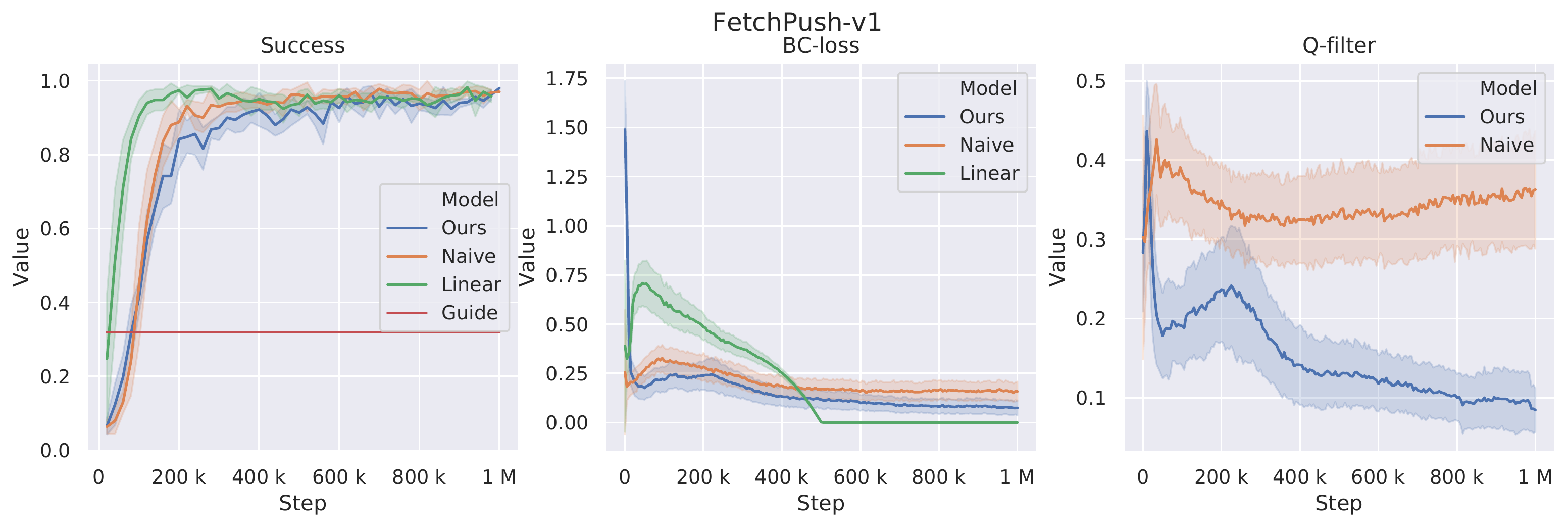}
    \end{subfigure}
    \begin{subfigure}[b]{\linewidth}
        \centering
        \includegraphics[scale=0.5]{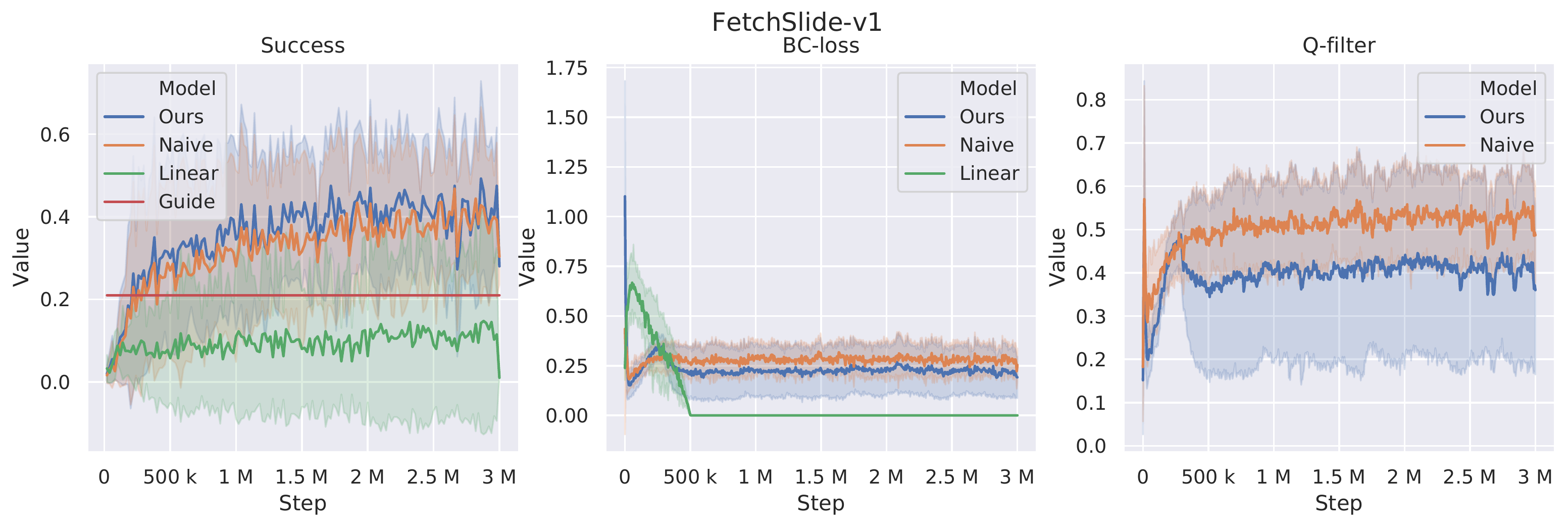}
    \end{subfigure}
    \caption{Results of experiments: The graphs show for each environment the success rate on the test set, the \gls{bc} loss, and the fraction of actions for which the indicator function evaluates to true in each minibatch. Note that the naive implementation of the Q-filter exhibit similar behaviour regardless of the environments difficulty and the pre-existing controllers proficiency, while our method is adaptive to these variables.}
    \label{fig:results}
\end{figure*}

\begin{figure}[thpb]
    \centering
    \begin{subfigure}[b]{0.48\linewidth}
        \centering
        \includegraphics[scale=0.35]{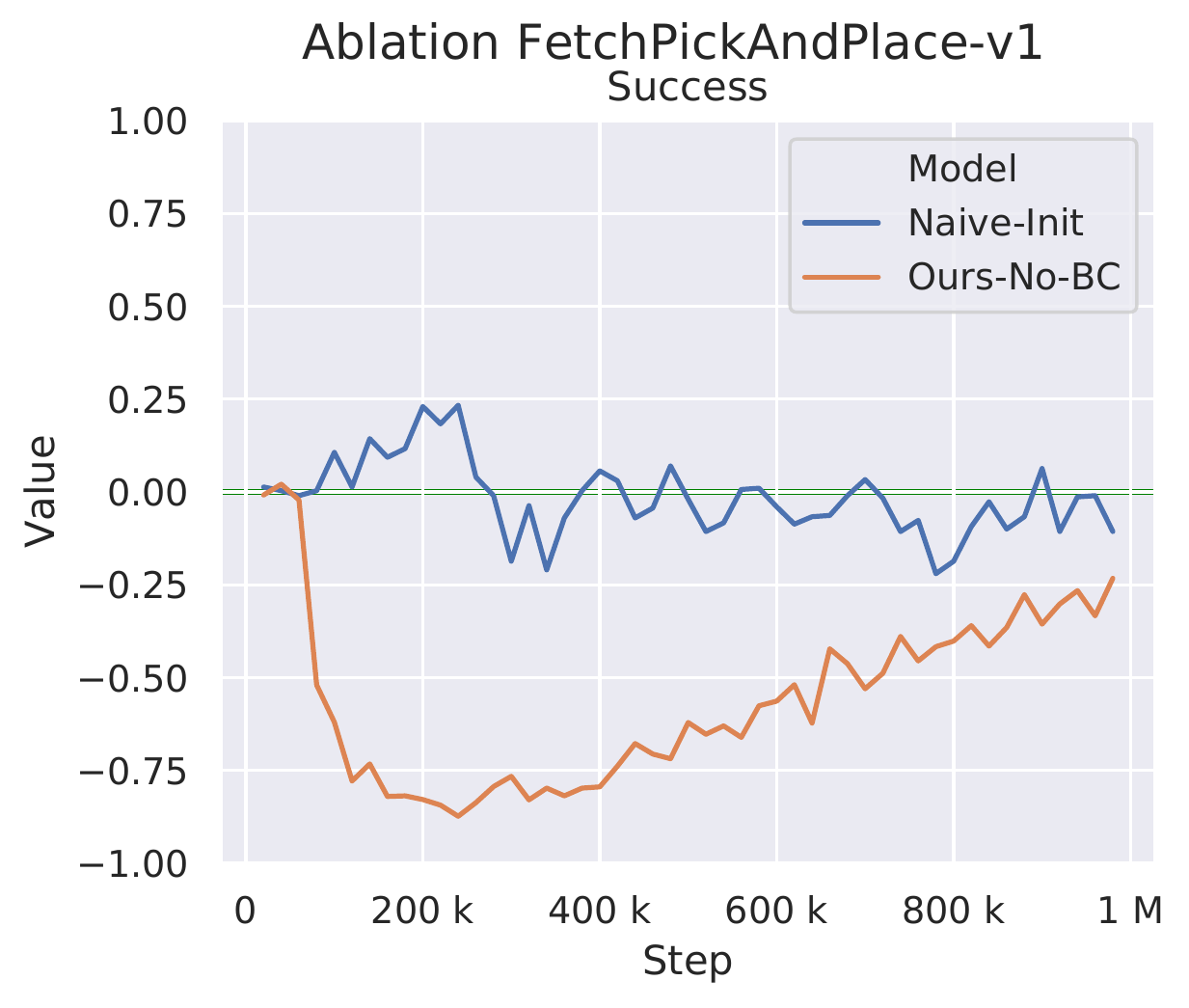}
    \end{subfigure}
    \begin{subfigure}[b]{0.48\linewidth}
        \centering
        \includegraphics[scale=0.35]{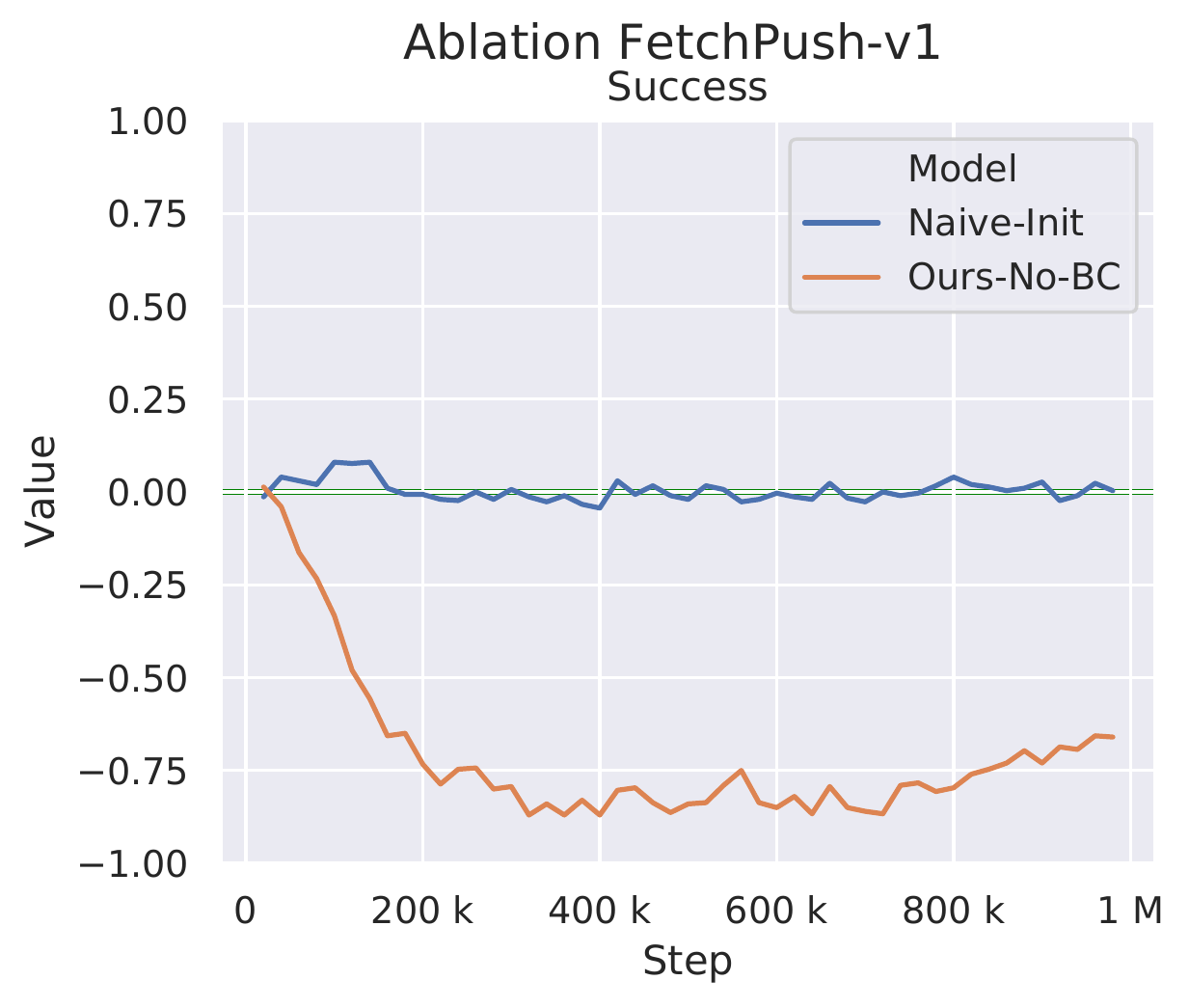}
    \end{subfigure}
    \begin{subfigure}[b]{\linewidth}
        \centering
        \includegraphics[scale=0.399]{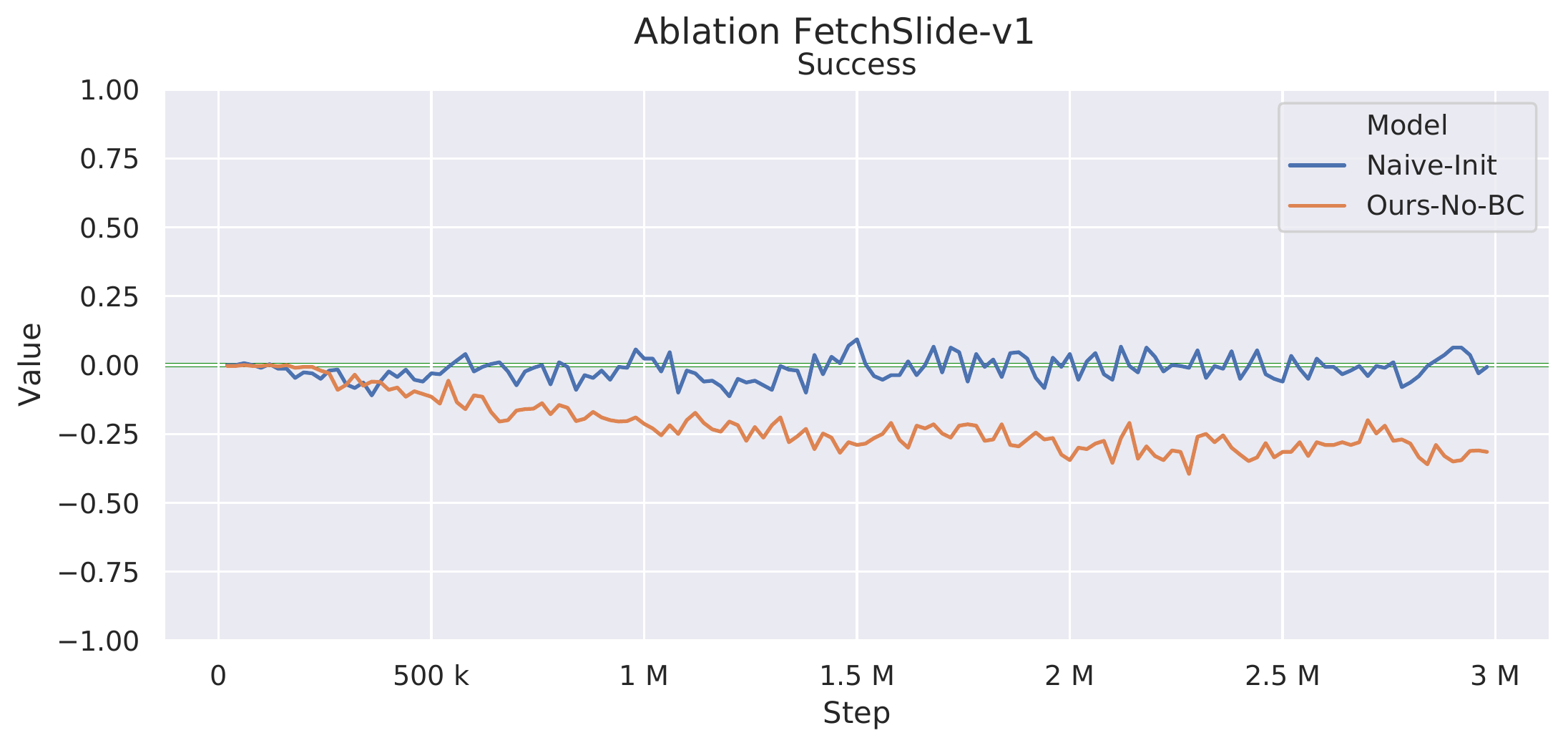}
    \end{subfigure}
    \caption{The graphs show the difference in performance when including initialization to $Q^G$ in the naive method and when removing the \gls{bc} loss term from our method, compared to its counterpart in Figure \ref{fig:results} as a baseline. The naive methods performance is largely unaffected by this modification, while the removal of the \gls{bc} loss from our method significantly decreases performance, down to a level comparable to having no guiding controller.}
    \label{fig:ablation}
\end{figure}

Figure \ref{fig:results} illustrates the unwanted behaviours of the naive Q-filter as described in section \ref{sec:method}: it typically starts low and converges to some value in the range of 0.3 - 0.5. Our method on the other hand delivers on its promises of adaptivity: it copies a large portion of the actions produced by the perfect guiding controller in the pick-and-place environment, selects only some of the actions produced by the pre-existing controllers in the slide and push environments, and decreases the frequency of imitation as the agent improves. The two Q-filter approaches have the same asymptotic performance for the push and slide environments, but our Q-filter method has considerable more success in the pick-and-place environment. The linear method matches the performance of the adaptive methods for the push environment, but falls a bit short in the pick-and-place environment, and is only able to find any success for a single seed in the most difficult slide task. The linear scheduling method seems to be in general the method with the quickest initial learning, due to indiscriminately imitating all actions from the guiding controller at the start, while the two Q-filter approaches seem to share a similar time at which improvements begin. 

\subsection{Evaluation of Results}
While our implementation of the Q-filter does lead to more deliberate application of the \gls{bc} loss, it does introduce a credit assignment problem. Since the two Q-functions that are compared correspond to two different controllers, it is unclear which actions in the action sequences generated by the controllers are responsible for the observed difference in value. Imitating only the initial action is therefore a potential source of error, but introducing some randomness and perturbation to the training process in \gls{rl} is a known measure to avoid local minima, and therefore this might not be a significant issue.

Even though the linear scheduling method is less complex than the other methods it can sometimes achieve similar or better performance in terms of takeoff time and asymptotic performance. The slide environment clearly illustrates that this approach require careful tuning, as the method is in this case unable to represent a policy capable of achieving any goals for most seeds. The Q-filters on the other hand are adaptive and provide reasonable performance for any task with little tuning, and in particular our method will imitate a good guiding controller to a larger degree than a bad guiding controller, unlike the linear scheduling method. Our version of the Q-filter does however require additional information in the form of $Q^E$, and the linear method might therefore be preferable in some scenarios despite its limitations.

Figure \ref{fig:ablation} shows the results of the ablation experiments. This experiment shows that initializing to $Q^G$ is not enough in and of itself to produce the results in Figure \ref{fig:results}, and that the \gls{bc} loss term is a vital part of the method. The knowledge contained in $Q^G$ is likely quickly lost through training on the limited data available in the beginning, and therefore does not significantly add to the performance. Figure \ref{fig:ablation} also shows that adding initialization to the naive version of the Q-filter does not alter the performance to a significant degree, further supporting the claim that the observed improvements in our method stem from an improved comparison in the Q-filter. 

The Q-function of the pre-existing controller need not be be obtained in a complicated manner such as Q-learning. Often, one has some historical data of usage of this controller, which one could use to estimate the expected sum of rewards in an entirely offline manner. Furthermore, one might not need a function approximator as complicated as a \gls{nn} for this purpose, and some simpler methods such as linear regression might be sufficient. This would remove the possibility of directly initializing the Q-function of the actor to that of the pre-existing controller, but one could simply pretrain the Q-network in a supervised manner on outputs from the pre-existing controllers Q-function to achieve much of the same effect. 

\section{CONCLUSION}\label{sec:conclusion}
We have shown that our method is capable of accelerating learning using guiding controllers with a wide spread of proficiency levels. An important further work is looking into how the optimality of the guiding controller affects the rate of convergence and asymptotic performance of the learning controller.

Having a pre-existing controller available online opens up many possibilities. If one knowns the stability properties of this controller, one can have the agent explore freely while still in the \gls{roa} of the guiding controller, and then have the guiding controller assume control and stabilize the system when necessary. In this way one can achieve a form of safe training, in the sense that only safe regions of the state space is visited, and risk of damage to the system is minimized. An interesting further work is a study on what fraction of actions the agent needs to control itself for learning to be successful in such a scenario.



\bibliography{references}

\end{document}